\documentclass[conference]{IEEEtran}
\IEEEoverridecommandlockouts
\usepackage{cite}
\usepackage{amsmath,amssymb,amsfonts}
\usepackage{algorithmic}
\usepackage{graphicx}
\usepackage{textcomp}
\usepackage{hyperref}
\usepackage{multirow}
\usepackage{xcolor}
\usepackage{subcaption}
\def\BibTeX{{\rm B\kern-.05em{\sc i\kern-.025em b}\kern-.08em
    T\kern-.1667em\lower.7ex\hbox{E}\kern-.125emX}}
\begin{document}

\title{Vision-Based Calorie Estimation for Bangladeshi Street Food: A Comparative Study of Detection and Regression Models
}
\author{
\IEEEauthorblockN{
\\ Aparup Dhar ,Antu Das, Pritom Barua, MD Tamim Hossain
}\

\IEEEauthorblockA{
\textit{Department of Computer Science \& Engineering} \\
Premier University, Chattogram, Bangladesh \\
\{iaparup, antudas468, pritombarua71\}@gmail.com, tamim.hossain@puc.ac.bd
}
}

\maketitle
\begin{abstract}
With obesity emerging as a major global health concern, accurate calorie estimation systems have become increasingly important for effective dietary management. Current vision-based approaches are inappropriate for Bangladeshi street food, which is widely consumed and culturally significant, because they mostly focus on Western cuisines and often overlook portion size. The purpose of this research is to offer a vision-based calorie estimation methodology that was created especially for street food in Bangladesh. Training, validation and testing splits were created from a proprietary dataset of 3,885 photos from six classes (Singara, Somusa, Puri, Peaju, Beguni, and Coin as a reference). Five detection and segmentation architectures, YOLOv8n, YOLO11n, YOLO12n, YOLO26n and RF-DETR, were methodically compared. Food dimensions were scaled using a Bangladeshi 5 Taka coin as a reference. To predict calories, extracted geometric characteristics were subsequently fed into machine learning regression models such as Random Forest, Gradient Boost, and AdaBoost. YOLO11n outperformed other models with the best detection performance, achieving 96.1\% mAP@50 and balanced mask metrics. With a mean absolute error (MAE) of 5.68, root mean squared error (RMSE) of 7.23, and an R2 score of 95.0\%, Random Forest regression produced the best results for calorie estimation. YOLO11n in conjunction with Random Forest regression offers a precise and effective calorie estimation for street food in Bangladesh, which is useful for dietary monitoring and mobile health apps. The code and dataset are publicly available on \href{https://github.com/hossain-tamim/BD-CAL}{https://github.com/hossain-tamim/BD-CAL} .
\end{abstract}


\section{Introduction}
Obesity is a major public health problem affecting a large portion of the world's population. It is a disease that is characterized by the accumulation of excessive or abnormal amounts of body fat, causing possible negative effects on the health of an individual. At a global scale, obesity rates have doubled since the year 1990, while as of 2022, approximately 890 million adults can be classified as obese \cite{who}. As such, obesity is now considered and regarded as a global epidemic that goes beyond traditional social and economic boundaries, as it disproportionately affects both lower and middle-income countries.

As a developing country, Bangladesh is currently facing this problem. The country was once synonymous with food insecurity, but now deals with a concurrent under-nutrition and over-nutrition situation. Reports from the Bangladesh Demographic and Health Survey (BDHS) found that 24\% of women and 16\% of men are overweight or obese, which is a staggering threefold increase since the year 2000 \cite{bangladesh_demographic}. In urbanized areas, where the majority of people lead a sedentary lifestyle and regularly consume street food on a daily basis, obesity rates surpass 30\% \cite{niport}. Processed snacks, fried foods, and sugary drinks now account for 40\% of urban calorie consumption \cite{fao}. The Centers for Disease Control and Prevention (CDC) recommends keeping track of food and beverage intake in order to ensure the amount of calories does not exceed the body's daily needs \cite{disease_control}.

Significant advancements have been made in estimating food calories through computer vision and deep learning. However, most solutions focus on Western cuisine and provide constant caloric values without considering portion sizes. To our knowledge, no prior work targets calorie estimation for Bangladeshi street food, which is widely consumed by the working-class population.

In this paper, we introduce a solution to accurately estimate the calorie content of Bangladeshi street food from a single image, accounting for actual food portion and weight by using a coin as a reference object to determine real-world food dimensions. We construct a dataset of the most popular Bangladeshi street foods, specifically built for this research, and use it to evaluate multiple state-of-the-art detection and segmentation architectures for this task.

\section{Literature Review}

\begin{figure*}[htbp]
    \centering
    \includegraphics[width=\linewidth,
    height=190px]{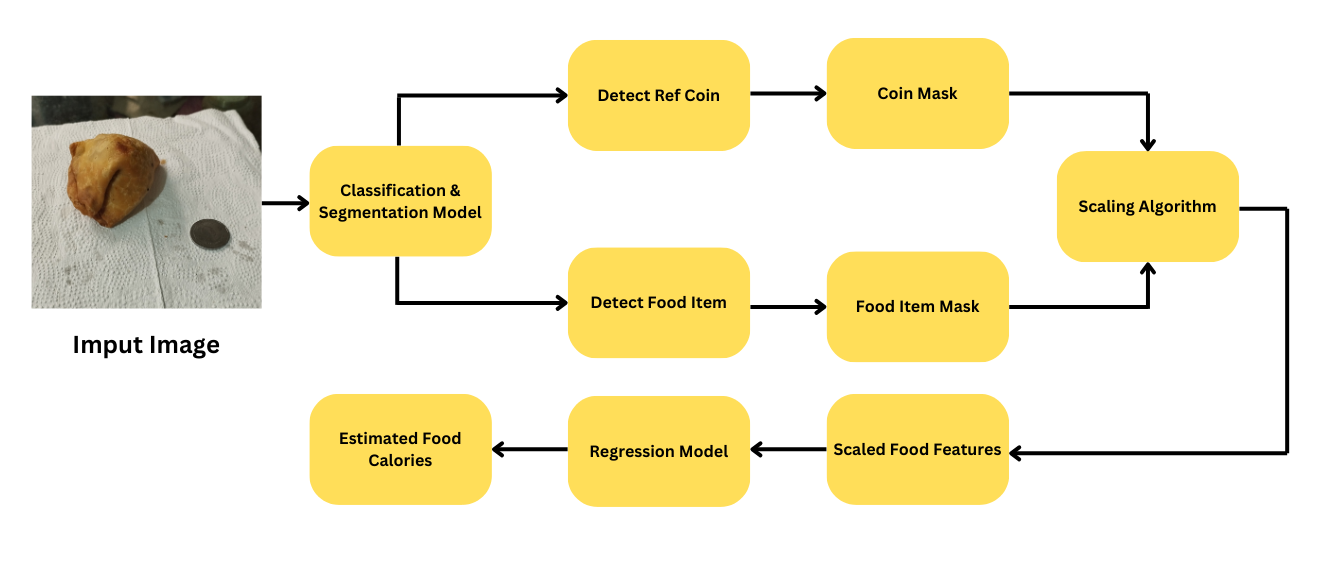}%
    \caption{Proposed System For Calorie Estimation}
    \label{proposed_system}
\end{figure*}

Recent years have seen extensive research on food calorie estimation, with many approaches leveraging convolutional neural networks (CNNs) for food detection and classification. Jiang et al. \cite{jiang2020deepfood} proposed a deep CNN using a Region Proposal Network from Faster R-CNN and VGGNet for feature extraction and classification. Although efficient due to a regression module for localization, their assumption of a constant 400g weight per item undermines accuracy due to variable portion sizes.

Mask R-CNN \cite{dollar2017mask} has also been widely adopted for object detection and segmentation in calorie estimation, with studies \cite{chiang2019food}, \cite{aditama2022indonesian} combining segmentation with linear regression for calorie prediction. However, these systems often depend on strict image capture protocols (e.g., fixed height), limiting usability.

To address scaling issues, Akpa et al. \cite{akpa2017smartphone} introduced chopsticks as a reference object to estimate food volume. Despite being practical, this limits applicability to East Asian cuisine and requires manual food classification. SCalE \cite{sadeq2018smartphone} estimated volume using distance-based calculations, pixel-to-centimeter ratios, and K-means clustering for segmentation. However, the system's reliance on user-specific measurements (e.g., arm length) and camera angles reduces flexibility and introduces potential errors.

A novel method by \cite{gao2019musefood} utilized a mobile device’s microphone and speaker with MLS signals for non-invasive vertical distance measurement. This two-image (top and side view) approach performed well but was sensitive to noise and required food to be in containers.

Steinbrener et al. \cite{steinbrener2023learning} proposed using monocular video and smartphone inertial data for metric volume estimation, combining Mask R-CNN outputs with LSTM-based processing. While achieving competitive accuracy, the method’s reliance on video input and lack of real-world robustness evaluation are noted limitations.

Finally, NIRSCam \cite{hu2022nirscam} introduced a mobile near-infrared sensing system for calorie estimation, outperforming image-based methods in accuracy and environmental robustness. It uses commercial LEDs for NIRS and three ML models to estimate macronutrients. However, the need for external hardware and high computational complexity may hinder widespread adoption.

\section{Methodology}

The proposed food calorie estimation system comprises three main components. First is the classification and segmentation module, which detects and identifies food items while segmenting them from the background. Next, the scaling algorithm adjusts the food item dimensions based on a reference object to determine real-world measurements. Finally, the calorie estimation model employs machine learning regression techniques to predict calorie values using the extracted features. Each component is detailed below.

\subsection{Classification \& Segmentation Model}
The proposed system accepts a single RGB image as input and
performs simultaneous food-item classification and instance
segmentation. Instance segmentation produces per-item binary masks
that are subsequently consumed by the scaling algorithm and the
regression pipeline for feature extraction.
 
To identify the best-performing detector for this task, we conduct
a systematic comparative study of five state-of-the-art
single-stage architectures: YOLOv8n,YOLO11n, YOLO12n,RF-DETR and YOLO26. All five models are trained from scratch on our custom Bangladeshi street-food dataset under identical hyper-parameter settings — 100 epochs, an input resolution of $640\times640$ pixels, and the AdamW optimiser with a cosine annealing learning-rate schedule — to ensure a fair,direct comparison of classification and instance segmentation capability.
 
\subsubsection{YOLOv8n}

The backbone is built on a refined CSPNet architecture that strengthens
gradient flow across stages, while a decoupled detection head
separates classification and regression tasks. As a single-stage
model, YOLOv8n completes all predictions in one forward pass,
making it substantially more efficient than two-stage detectors
such as Mask~R-CNN.

\subsubsection{YOLO11n}

YOLO11n is the next-generation Ultralytics architecture released
in September 2024.By decoupling the classification and localisation
branches, YOLO11n improves recognition accuracy for visually similar food classes — a common challenge in the Bangladeshi street-food domain.
 
\subsubsection{YOLO12n}

YOLO12n, presented at NeurIPS 2025, adopts an
attention-centric design philosophy within the YOLO
framework.These architectural choices make YOLO12n particularly effective for distinguishing food items that share similar textures or colours but differ in shape — a recurring challenge in our street-food
imagery.
 
\subsubsection{RF-DETR}

RF-DETR uses weight-sharing Neural Architecture Search (NAS) to jointly
optimise accuracy and latency, and achieves state-of-the-art performance on both COCO and the Roboflow100-VL benchmark. We
include RF-DETR in our evaluation to assess whether transformer-based backbones offer complementary strengths — particularly in generalising to fine-grained, domain-specific food categories with limited training data.
 
\subsubsection{YOLO26n}

It is the first Ultralytics detector to be fully end-to-end: Distribution Focal Loss (DFL) is removed to streamline bounding-box regression and improve export compatibility, and the NMS-free prediction head eliminates post-processing latency.  A single unified pipeline supports detection, instance segmentation, pose estimation, oriented bounding-box detection, and classification across five size
scales, making YOLO26n the most versatile and deployment-friendly
architecture evaluated in this work.
 
\subsubsection{Model Selection and Integration}

The best-performing architecture is selected based on mAP@0.5
and mAP@0.5:0.95 on our held-out test split. The binary masks
produced by the chosen model are passed to the scaling algorithm,
which converts pixel-space dimensions into real-world measurements
using a Bangladeshi 5~Taka coin (diameter 25.5~mm) as a reference
object. The resulting scaled geometric features — height, width,
area, perimeter, and class label — are then forwarded to the
regression model for calorie prediction, as described in the
following subsections.

\subsection{Scaling Algorithm}

Accurate food calorie estimation requires determining the real-world dimensions of food, but images captured from different angles and distances can distort these measurements. To address this, our system uses a reference object of known size to compute a scaling factor that converts image measurements into real-world units. We propose a standard coin as the reference, chosen over alternatives like teaspoons, photographic scales, or hands, due to its accessibility, consistency, and ease of use. In our implementation, a Bangladeshi 5 Taka coin (25.5 mm diameter) is detected in the image, its pixel size is measured, and a scaling factor is derived. This factor is then applied to segmented food items to obtain real-world dimensions, as detailed in the next section.

\[
\begin{aligned}
    \text{Height of Food Item} &= F_h \text{ px} \nonumber \\ 
    \text{Width of Food Item} &= F_w \text{ px} \nonumber \\
    \text{Diameter of Ref Coin} &= D_c = 25.5 \text{ mm} \nonumber
\end{aligned}
\]
Scale factors for height and width are given by:

\begin{equation}
\text{Scale Factor (Height), } S_h = \frac{D_c}{F_h} \label{height}
\end{equation}
\begin{equation}
\text{Scale Factor (Width), } S_w = \frac{D_c}{F_w} \label{width}
\end{equation}

Thus, the overall scale factor is:

\begin{equation}
    S_f = \frac{S_h + S_w}{2} \label{sf}
\end{equation}

\subsection{Estimation / Regression Model}
The final step in the system is calorie estimation, which is performed using a machine learning regression model. Once the binary mask of the food item is obtained from the segmentation model, we extract key features such as height, width, area, perimeter, and class label using OpenCV's contour feature functions \cite{contour}. These features are then scaled using the previously calculated scaling factor to reflect real-world dimensions. The scaled features are organized into a structured data frame and passed into the regression model, which outputs a continuous value representing the estimated calorie content of the food item.

\section{Dataset Description}

For the proposed system, we require two datasets. One for training the classification, segmentation model, and a separate dataset with food items alongside a reference object coin for constructing our regression dataset. Both datasets are separately discussed in detail below.

\subsection{Classification/Segmentation Dataset}
Although many datasets exist for Western and Asian foods, to our knowledge, there is no dataset focused on Bangladeshi cuisine that includes instance segmentation masks alongside a reference object. To address this, we created a dataset based on popular Bangladeshi street foods \cite{choudhury2021identification}, \cite{islam2017street}, comprising six classes: Singara, Somusa, Puri, Peaju, Beguni, and a reference object class for the coin. Images were collected both from the web and onsite, ensuring diversity. Web images offer a wide variety of environments and contexts, while physical data captures real-world challenges like poor lighting, occlusions, and unusual angles, improving the generalization ability of the segmentation and classification model.

\begin{figure}[htbp]
    \centering
    \subfloat[]{%
        \includegraphics[width=\linewidth]{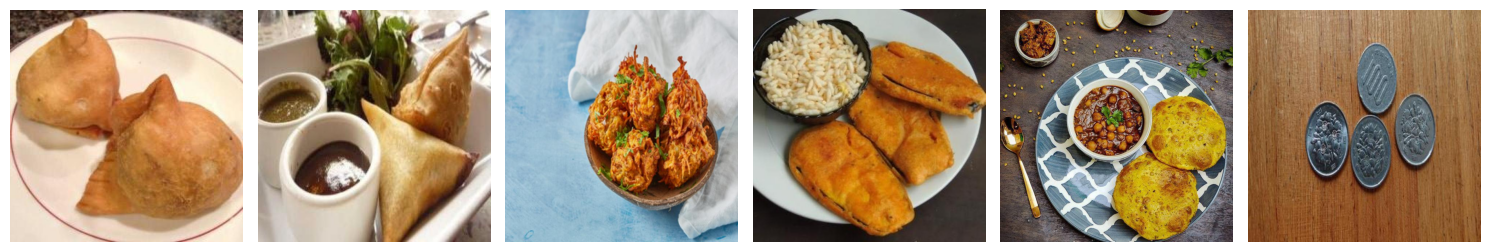}%
    }\\
    \subfloat[]{%
        \includegraphics[width=\linewidth]{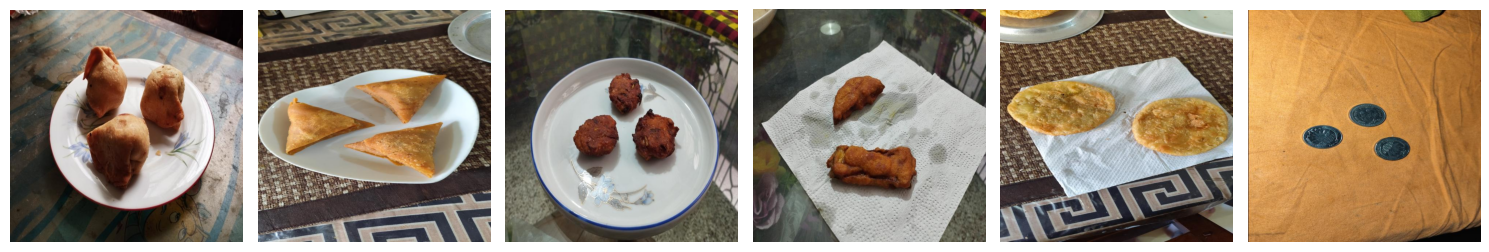}%
    }
    
    \caption{Sample Images From Dataset (a) Images Collected From Web (b) Images Collected Onsite }
    \label{sample_dataset}
\end{figure}

The base dataset contains 1,847 images, which were expanded through augmentation to 3,885 images. These are split 80\% for training (3,146 images), 10\% for validation (370 images), and 10\% for testing (369 images). The total number of instances per class is shown in Table~\ref{class_instance}, and sample images are presented in Fig.~\ref{sample_dataset}

\begin{table}[htbp]
\caption{Class Instance Distribution}\label{class_instance}%
\setlength{\tabcolsep}{18pt}
\begin{center}
\begin{tabular}{|c|c|c|c|}
\hline
\textbf{Class} & \textbf{Train} & \textbf{Test} & \textbf{Valid} \\
\hline
Singara & 467 & 172 & 136 \\
\hline
Somusa  & 567 & 185 & 173 \\
\hline
Puri    & 415 & 149 & 144 \\
\hline
Peaju   & 764 & 271 & 287 \\
\hline
Beguni  & 603 & 207 & 203 \\
\hline
Coin    & 469 & 162 & 161 \\
\hline
\end{tabular}
\end{center}
\end{table}

\subsection{Estimation / Regression Dataset}
For the regression dataset, we prepared a separate set of images that were passed through the trained model to extract features like height, width, area, perimeter, and class. The dependent variable, calories, is calculated by multiplying the food’s weight (measured in grams with a digital scale) by its caloric density. The calorie calculation formula is shown below.

\begin{equation}
C = W \times D \label{calorie}
\end{equation}

where:  
\begin{align}
C &= \text{Calories} \nonumber \\
W &= \text{Weight of food (grams)} \nonumber \\
D &= \text{Caloric density of food (kcal per gram)} \nonumber
\end{align}

The caloric density of every food item used in the research is determined according to an online nutritional database \cite{Nutritionix}. The specific values are shown in Table~\ref{calorie_density}  below. Caloric density has been considered for standard variants of each food item. For example, in the case of Singara, commonly used ingredients such as potato and peas have been considered as a baseline. Alternative variants using meat or other ingredients have not been considered
in this analysis.

\begin{table}[htbp]
\caption{Calorie Density Values By Class}\label{calorie_density}%
\setlength{\tabcolsep}{18pt}
\begin{center}
\begin{tabular}{|c|c|}
\hline
\textbf{Class} & \textbf{Calorie Density} \\
\hline 
Singara & 2.61 \\ \hline 
Somusa  & 2.11 \\ \hline 
Puri    & 2.44 \\ \hline 
Peaju   & 1.18 \\ \hline 
Beguni  & 1.48 \\ \hline 

\end{tabular}
\end{center}
\end{table}

\begin{figure}[htbp]
    \centering
    \includegraphics[width=\linewidth,]{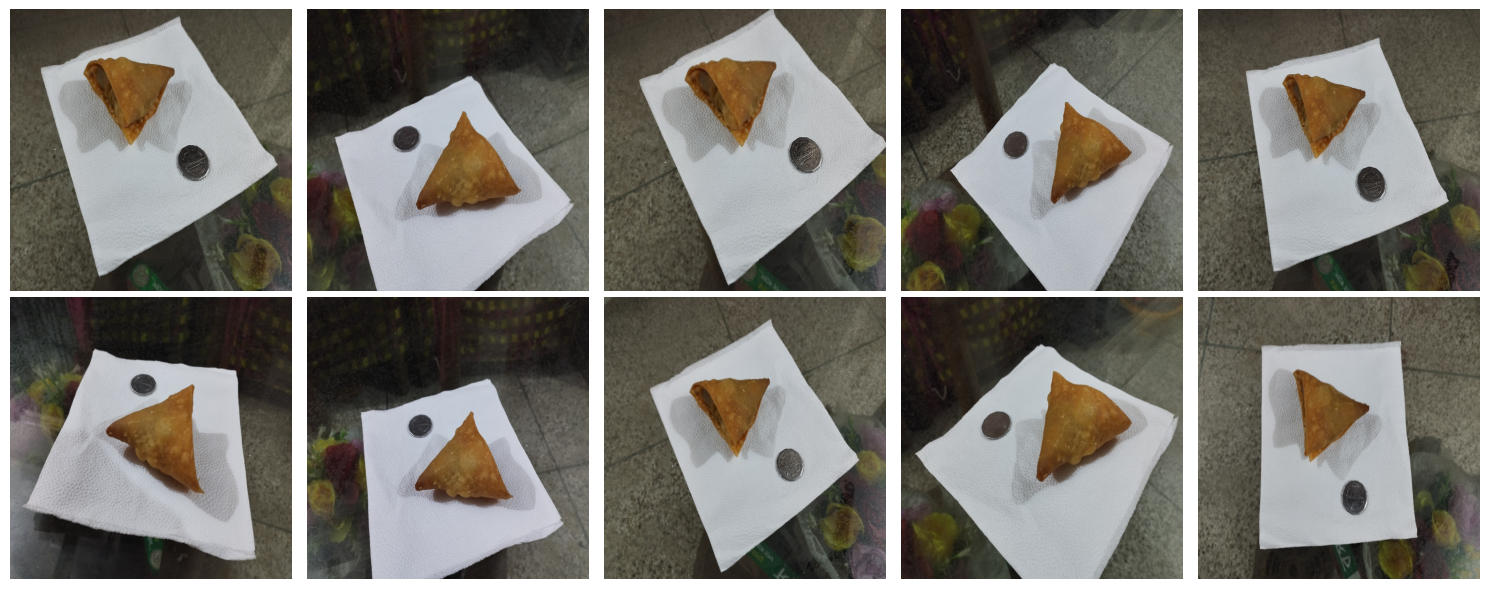}
    \caption{Example of Multiple Images Taken From Different Angles of a Singular Food Item}
    \label{singular_food}
\end{figure}

Each food item was placed beside the reference coin and photographed from 10 different angles to capture visual variations. Although the appearance changes, the calorie content remains constant. Fig.~\ref{singular_food} illustrates example images of a single food item from multiple views.

After passing each image through our scaling algorithm, we obtain the features, which are appended with caloric values in a comma-separated values (CSV) file. After preprocessing, we obtain our regression dataset comprising 644 records. The distribution of classes in the regression dataset is shown below in Table~\ref{sample_image}.

\begin{table}[h]
\caption{Sample Image Distribution}\label{sample_image}%
\setlength{\tabcolsep}{18pt} 
\begin{center}
\begin{tabular}{|c|c|}
\hline
\textbf{Class} & \textbf{Records} \\
\hline
Singara & 128 \\ \hline
Somusa  & 131 \\ \hline 
Puri    & 116 \\ \hline 
Peaju   & 133 \\ \hline
Beguni  & 136 \\
\hline
\end{tabular}
\end{center}
\end{table}

\section{Preprocessing Steps}
Image preprocessing enhances quality and standardizes input for deep learning. To improve generalization and mitigate overfitting, training data is augmented using horizontal flips and 90° rotations, increasing the number of training samples from 1,847 to 3,885 images, of which 3,146 were used for training. Additionally, all images are resized to 640×640 pixels to ensure a consistent input shape, reduce computational load, and avoid feature distortion.

For regression modeling, the data undergoes further preprocessing. Categorical class names are converted into a numeric format using one-hot encoding, where each class is represented as a binary vector. Feature scaling is applied through min-max normalization to ensure that features with larger magnitudes, such as area or perimeter, do not dominate the learning process. Finally, Z-score normalization is used to eliminate outliers from mispredictions or segmentation errors, with a threshold set at 2.

\section{Performance Metrics}

For Bangladeshi street food calorie estimation, we categorize performance metrics into classification/segmentation and regression. We use Precision, Recall, and mean Average Precision (mAP) for detection and segmentation accuracy, along with GFLOPs to measure computational cost. Calorie estimation is evaluated using MAE, RMSE, and the R² score to indicate variance explained by the model.

\section{Results}

\begin{table*}[htbp]
\caption{Performance Metrics Comparison}\label{tab1}%
\footnotesize
\setlength{\tabcolsep}{6pt} 
\renewcommand{\arraystretch}{1.25} 
\begin{center}
\begin{tabular}{|c|c|c|c|c|c|c|c|c|}
\hline

\multirow{2}{*}{\textbf{Model}}&\multicolumn{4}{|c|}{\textbf{Bounding Box}}
&\multicolumn{4}{|c|}{\textbf{Mask}} \\
\cline{2-9}
 & \textbf{P} & \textbf{R} & \textbf{mAP50} & \textbf{mAP50-95} & \textbf{P} & \textbf{R} & \textbf{mAP50} & \textbf{mAP50-95} \\

\hline
Mask R-CNN (ResNet50, C4) & 90.1\% & 81.6\% & 90.1\% & 76.7\% & 89.4\% & 79.2\% & 89.3\% & 74.4\% \\ \hline 
Mask R-CNN (ResNet50, DC5) & 91.2\% & 82.4\% & 91.1\% & 78.2\% & 91.2\% & 81.9\% & 91.2\% & 77.7\% \\ \hline 
Mask R-CNN (ResNet50, FPN) & 90.5\% & 81.7\% & 90.5\% & 77.6\% & 90.8\% & 82.4\% & 90.8\% & 78.5\% \\ \hline 
YOLOv8n & 93.5\% & 89.2\% & 93.1\% & 86.2\% & \textbf{91.7\%} & 87.4\% & 91.6\% & \textbf{81.1\%} \\ \hline 
\textbf{YOLO11n} & \textbf{94.1\%} & \textbf{91.3\%} & \textbf{96.1\%} & \textbf{86.8\% }& 90.6\% & \textbf{93\%} & \textbf{94.9\%} & 80.4\% \\ \hline 
YOLO12n & 82.3\% & 78.8\% & 87.5\% & 70.9\% & 85.7\% & 74.8\% & 85.4\% & 68\% \\ \hline
YOLO26n & 84.3\% & 82.4\% & 88.9\% & 75.9\% & 84.7\% & 81.8\% & 87.4\% & 70.3\% \\ \hline
RF-DETR & 67.3\% & 71.8\% & 75\% & 59\% & 70.7\% & 77.8\% & 72\% & 70\% \\ \hline
\end{tabular}
\end{center}
\end{table*}

Table~\ref{tab1} reveals YOLO11n’s superior performed across all detection metrics. YOLO11n achieved 94.1\% precision, 91.3\% recall, and 96.1\% mAP@50, outperforming YOLOv8n and other baselines. Mask metrics also confirmed YOLO11n’s strength, with 94.9\% mAP@50 and balanced recall across classes. For example, Beguni improved from 86.5\% (YOLOv8n) to 88.8\% (YOLO11n), and Peaju from 87.7\% to 89.8\%. These consistent gained demonstrate YOLO11n’s robustness across diverse food categories.

\subsection{Confusion Matrix}

Beguni, Coin, Peaju, Puri, Singara, Somusa, and background were the seven categories in which the confusion matrix offers a thorough summary of categorisation performance. Misclassifications were shown by off-diagonal values, whereas correctly classified occurrences are represented by diagonal entries. With 237 accurate predictions—the most of any class—the model had a remarkable recognition ability for Peaju. With 181 and 159 accurate predictions, respectively, Beguni and Coin also demonstrate strong detection. Somusa made 164 accurate predictions, demonstrating dependable detection, while Puri and Singara performed satisfactorily with 130 and 123 correctly recognised samples.

\begin{figure}[htbp]
    \centering
    \includegraphics[width=\linewidth,]{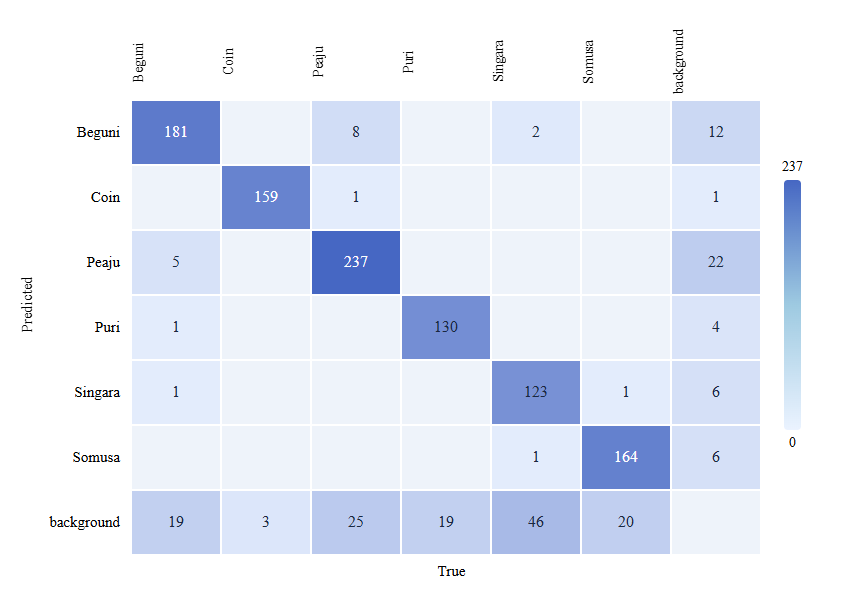}
    \caption{Confusion Matrix for test dataset  }
    \label{confusion_matrix}
\end{figure}


\subsection{Computational Complexity Comparison}

\begin{table}[htbp]
\caption{Comparison of Models by Computational Complexity and Size}
\label{tab:model_comparison}
\renewcommand{\arraystretch}{1.2} 
\setlength{\tabcolsep}{4pt} 
\begin{center}
\begin{tabular}{|c|c|c|c|c|}
\hline
\textbf{Model} & \textbf{GFLOPs} & \shortstack{\textbf{Params} \\ \textbf{(M)}} & \shortstack{\textbf{Time} \\ \textbf{(ms)}} & \shortstack{\textbf{Size} \\ \textbf{(MB)}} \\
\hline 
M-RCNN (R50, C4)         & 822   & 34   & 248.24 & 420.51 \\ \hline 
M-RCNN (R50, DC5)        & 344   & 166  & 149.83 & 2060.00 \\ \hline 
M-RCNN (R50, FPN)        & 447   & 44   & 58.16  & 526.23 \\ \hline 
YOLOv8n                 & 12.0  & 3.2  & 3.8    & 6.80 \\ \hline
YOLO26n                 & 10.2  & 3.05  & 7.22 & 13   \\ \hline
YOLO12n                 & 9.8   & 2.82  & 8.32 & 12   \\ \hline
YOLO11n                 & 9.7   & 2.85  & 16.24 & 12 \\ \hline
RF-DETR                 & 9.5   & 2.80  & 15.45 & 12 \\ \hline

\label{tab:complexity}
\end{tabular}
\end{center}
\end{table}

All models are compared in terms of GFLOPs, parameters, inference time, and model size in Table~\ref{tab:complexity}. Mask~R-CNN variations are not viable for real-time application because to their computational cost, which ranges from 344 to 822~GFLOPs, inference durations up to 248~ms, and model sizes surpassing 420~MB.The efficiency of all five single-stage detectors is significantly higher. YOLO26n further decreases complexity to 10.2~GFLOPs and 3.05M parameters with NMS-free end-to-end inference, providing up to 43\% faster CPU throughput than previous YOLO generations. YOLOv8n delivers 12.0~GFLOPs at 3.8~ms inference with just a 6.80~MB. The lightest models tested, YOLO12n and YOLO11n, each weigh less than 9.8GFLOPs and 12MB. RF-DETR, although slower on CPU (71.22ms) because of its deformable attention decoder, provides better domain generalisation through its DINOv2 backbone, which makes it ideal for small, domain-specific datasets like ours.

\subsection{Regression / Estimation Model Results}
\begin{table}[htbp]
\caption{Performance Metrics of Machine Learning Models}
\label{tab:model_metrics}
\begin{center}
\begin{tabular}{|c|c|c|c|}
\hline 
\textbf{Model} & \textbf{MAE} & \textbf{MSE} & \textbf{R² Score} \\
\hline 
Linear Regression   & 13.02  & 382.40 & 75.0\% \\ \hline 
KNN                 & 10.32   & 1081.88 & 26.0\% \\ \hline 
Decision Tree      & 0.30   & 7.21 & 92.0\% \\ \hline 
\textbf{Random Forest}       & \textbf{5.68}   & \textbf{121.10} & \textbf{95.0\%} \\ \hline 
Gradient Boost      & 7.84   & 117.24 & 92.0\% \\ \hline 
AdaBoost            & 8.10   & 102.15 & 92.0\% \\
\hline 
\end{tabular}
\end{center}
\end{table}

The performance evaluation of various machine learning models for calorie estimation showed that ensemble and non-linear models outperform traditional linear regression. Random Forest stood out with the lowest MAE of 5.68, RMSE of 7.23, and the highest R² of 95\%, indicating accurate predictions with minimal errors. This highlights the advantage of ensemble learning in captured complex patterns and reducing overfitting. Gradient Boost and AdaBoost also performed well, each achieving around 92\% R², demonstrating their ability to model underlying data relationships.

\section{Statistical Validation}

We performed a k-fold (k = 5) evaluation on the dataset to determine the statistical significance of performance differences between YOLOv8n and YOLO11n. While the YOLO11n model had a greater mean accuracy of 91.87\% (SD = 3.33), the YOLOv8n model recorded a mean accuracy of 90.74\% (SD = 2.88). Table ~\ref{tab:statis} provides a summary of these findings.To ascertain whether the observed performance difference between the two models was statistically significant, a paired t-test was used. YOLO11n consistently outperformed YOLOv8n across folds, as indicated by the test's positive t-statistic. Though practically substantial, the improvement was not statistically significant under the current experimental configuration, as indicated by the p-value exceeding the traditional 0.05 threshold.Additionally, both models showed similar variability, with YOLO11n exhibiting a slightly greater standard deviation (SD = 3.33) than YOLOv8n (SD = 2.88), indicating that YOLO11n's performance varied more across folds. These results show that YOLO11n is the better option overall, but more research using bigger datasets and more precise hyperparameter tuning is required to demonstrate statistical resilience.

\begin{table}[htbp]
\caption{  Statistical validation results comparing YOLOv8n and YOLO11n accuracies across 5 
folds}
\label{tab:statis}
\begin{center}
\begin{tabular}{|c|c|c|c|}
\hline 
\textbf{Fold} & \textbf{YOLOv8n} & \textbf{YOLO11n}  \\
\hline 
1   & 93.50  & 94.50 \\ \hline 
2                 & 89.20   & 91.24 \\ \hline 
3     & 93.10   & 96.10 \\ \hline 
4     & 86.20   & 86.80 \\ \hline
5     & 91.70   & 90.70 \\ \hline

Mean      & 90.74   & 91.87 \\ \hline 
SD            & 2.88   & 3.33 \\
\hline 
\end{tabular}
\end{center}
\end{table}

\section{Conclusion and Future Work}

This paper presents a novel approach for estimating calories in Bangladeshi street food, addressing the lack of region-specific nutritional data. We compiled a dataset of 3,885 images of popular street foods, ensuring real-world relevance.  For calorie prediction, we use a regression-based method that reduces error margins, making it practical for applications like mobile apps.

Looking ahead, we aim to scale food items accurately without reference objects, exploring techniques like depth estimation from images. While our dataset includes many popular dishes, it needs expansion to cover more traditional foods. Additionally, optimizing the model for efficiency is crucial for performance on low-end mobile devices, enhancing accessibility, especially in areas with limited technology.

\bibliographystyle{ieeetr}
\bibliography{references}
\end{document}